\documentclass{llncs}

\usepackage{graphicx} 
\usepackage{amssymb} 
\usepackage{algorithmic,algorithm}

\newcommand{\Real}{\mathop{\rm I\kern-.2emR}}
\newcommand{\V}{{\cal N}}

\begin{document}

\title{NILS: a Neutrality-based Iterated Local Search\\and its application to Flowshop Scheduling}
\titlerunning{Neutrality-based Iterated Local Search for Flowshop Scheduling}

\author{Marie-El\'eonore Marmion\inst{1,2} \and Clarisse~Dhaenens\inst{1,2} \and Laetitia Jourdan\inst{2} \and \\ Arnaud Liefooghe\inst{1,2} \and S\'ebastien Verel\inst{2,3}}

\authorrunning{M.-E. Marmion, C. Dhaenens, L. Jourdan, A. Liefooghe and S. Verel}

\institute{Universit\'e Lille 1, LIFL -- CNRS, France
\and INRIA Lille-Nord Europe, France
\and University of Nice Sophia Antipolis -- CNRS, France \\
\email{firstname.lastname@inria.fr}
}

\maketitle

\begin{abstract}
This paper presents a new methodology that exploits specific characteristics from the fitness landscape.
In particular, we are interested in the property of \emph{neutrality},
that deals with the fact that the same fitness value is assigned to numerous solutions from the search space.
Many combinatorial optimization problems share this property,
that is generally very inhibiting for local search algorithms.
A neutrality-based iterated local search, that allows neutral walks to move on the plateaus,
is proposed and experimented on a permutation flowshop scheduling problem with the aim of minimizing the makespan.
Our experiments show that the proposed approach is able to find improving solutions compared with a classical iterated local search.
Moreover, the tradeoff between the exploitation of neutrality and the exploration of new parts of the search space is deeply analyzed.
\end{abstract}

\section{Motivations}

Many problems from combinatorial optimization, and in particular from scheduling, present a high degree of neutrality. 
Such a property means that a lot of different solutions have the same fitness value. 
This is a critical situation for local search techniques (although this is not the only one),
since it becomes difficult to find a way to reach optimal solutions. 
However, up to now, this neutrality property has been under-exploited to design efficient search methods.
Barnett~\cite{barnett01netcrawling} proposes a heuristic (the {\it Netcrawler process}), adapted to neutral landscapes, 
that consists of a random neutral walk with a mutation mode adapted to local neutrality.
The per-sequence mutation rate is optimized to jump from one neutral network to another.
Stewart \cite{stewart2001} proposes an {\it Extrema Selection} for evolutionary optimization
in order to find good solutions in a neutral search space. 
The selection aims at accelerating the evolution during the search process once most solutions from the population have reached the same level of performance.
To each solution is assigned an endogenous performance during the selection step to explore the search space area
with the same performance more largely, and to reach solutions with better fitness values.
Verel et al. \cite{VEREL:2004:HAL-00160035:1} propose a new approach, the \textit{Scuba search}, which has been tested on Max-SAT problems and on NK-landscapes with neutrality. This local search exploits the evolvability (ability of random variations to reach improving solutions) of neutral networks. On neutral plateaus, the search is guided by the evolvability of solutions. When there is no neighboring solution with a higher evolvability on the plateau, a fitness ``jump'' is performed.

Knowing that the permutation flowshop scheduling problem (PFSP) of minimizing the makespan is
$\cal{NP}$-hard in the general case~\cite{johnson1954},
a large number of metaheuristics have been proposed so far for its resolution. 
Recently, 25 methods have been tested and their performance has been compared by Ruiz and Maroto~\cite{ruiz:2005}.
The Iterated Local Search of St{\" u}tzle~\cite{stutzle:1998} has been shown to be one of the most efficient algorithm to solve Taillard's FSP instances \cite{taillard:1993}.
More recently, Ruiz and St{\"u}tzle \cite{ruiz_stutzle:2006} proposed a simple variant of St{\" u}tzle's ILS based on greedy mechanisms.
The perturbation is made of a destruction and of a construction phase: 
jobs are removed from the solution and then re-inserted in order to get a new configuration that yields the best possible fitness value.
Given that the PFSP is known to have a high neutrality with respect to the insertion neighborhood operator \cite{Marmion:Lion2011},
we can assume that a lot of moves which do not change the fitness value are allowed with such a perturbation strategy.
Furthermore, the acceptance criterion used in \cite{ruiz_stutzle:2006} has the particularity to accept equivalent, and then neutral, solutions.
We argue that this neutrality property could be used more explicitly, and in a more simple way.

This paper will not bring to a new heuristic which produces some new best-known solutions for a given scheduling problem.
The goal of this preliminary work is to give a minimal, and yet efficient approach based on Iterated Local Search (ILS),
which exploits the neutrality property of the search space in a novel way.
The approach explicitly balances the search between the exploitation of the plateaus in the landscape,
and the exploration of new parts of the search space.
Two main questions are addressed in this paper:
($i$) What are the performances of this neutrality-based approach on solving a difficult scheduling problem where neutrality arises?
($ii$) What are the costs and the benefits of such an exploitation?
A Neutrality-based Iterated Local Search (NILS) that performs neutral walks along the search is proposed. 
The performances and the dynamics of NILS are deeply analyzed on the PFSP.

The paper is organized as follows.
Section~\ref{sec:background} is dedicated to the flowshop scheduling problem,
to the required definitions of neutral fitness landscapes,
and to the main principles of iterated local search.
Section~\ref{sec:nils} presents the Neutrality-based Iterated Local Search (NILS) proposed in the paper.
Section~\ref{sec:analysis} is devoted to the analysis of the NILS efficiency to solve the PFSP.
Finally, the last section concludes the paper and gives suggestions for further research.

\section{Background}
\label{sec:background}

\subsection{The Permutation Flowshop Scheduling Problem}
The Flowshop Scheduling Problem is one of the most investigated scheduling problem from the literature.
The problem consists in scheduling $N$~jobs $\{J_1,J_2,\dots,J_N\}$ on $M$ machines $\{M_1,M_2,\dots,M_M\}$.
Machines are critical resources, {\itshape i.e.} two jobs cannot be assigned to the same machine at the same time.
A job $J_i$ is composed of $M$~tasks $\{t_{i1},t_{i2},\dots,t_{iM}\}$, 
where~$t_{ij}$ is the $j^{th}$ task of~$J_i$, requiring machine~$M_j$.
A processing time~$p_{ij}$ is associated with each task~$t_{ij}$.
We here focus on a Permutation Flowshop Scheduling Problem (PFSP), 
where the operating sequences of the jobs are identical for every machine.
As a consequence, a feasible solution can be represented by a permutation $\pi_N$ of size $N$
(determining the position of each job within the ordered sequence),
and the size of the search space is then $|S| = N!$.

In this study, we consider the minimization of the makespan, {\itshape i.e.}~the total completion time, as the objective function.
Let  $C_{ij}$ be the completion date of task~$t_{ij}$, the makespan can be defined as follows: $C_{max} = \max_{i \in \{1, \dots, N\}} \{C_{iM}\}$

\subsection{Neighborhood and Local Search}
\label{subsec:nbh}
The design of local search algorithms requires a proper definition of the neighborhood structure for the problem under consideration.
A \emph{neighborhood structure} is a mapping function $\mathcal{N}: S \rightarrow 2^S$ 
that assigns a set of solutions $\mathcal{N}(s) \subset S$ to any feasible solution $s \in S$.
$\mathcal{N}(s)$ is called the \emph{neighborhood} of $s$, 
and a solution $s' \in \mathcal{N}(s)$ is called a \emph{neighbor} of $s$.
A neighbor of solution $s$ results of the application of a \emph{move operator} performing a small perturbation to solution $s$.
This neighborhood is one of a key issue for the local search efficiency.
A solution $s^{*}$ is a \emph{local optimum} iff no neighbor has a better fitness value:
$\forall s \in \V(s^{*})$, $f(s^{*}) \leq f(s)$. 

For the PFSP, we consider the \emph{insertion operator}.
This operator is known to be one of the best-performing neighborhood structure for the PFSP~\cite{stutzle:1998,ruiz2005}.
It can be defined as follows.
Let us consider an arbitrary solution, represented here by a permutation of size $N$ (the number of jobs).
A job located at position $i$ is inserted at position~$j \neq i$.
The jobs located between positions~$i$ and~$j$ are shifted.
The number of neighbors {\itshape per} solution is then $(N-1)^2$~\cite{stutzle:1998}, where $N$ stands for the size of the permutation.

\subsection{Neutral Fitness Landscape}

A fitness landscape \cite{stadler:1995} can be defined by a triplet $(S, \V, f)$,
where $S$ is a set of admissible solutions ({\itshape i.e.} the search space), $\V: S \longrightarrow 2^{S}$, a neighborhood structure, 
 and $f: S \longrightarrow \Real$ is a fitness function that can be pictured as the \textit{height} of the corresponding solutions.

A \emph{neutral neighbor} is a neighboring solution having the same fitness value,
and the set of neutral neighbors of a solution $s \in S$ is then $\V_n(s) = \{ s' \in \V(s) ~|~ f(s^\prime) = f(s) \}$.
The \emph{neutral degree} of a given solution is the number of neutral solutions in its neighborhood.
A fitness landscape is said to be \emph{neutral} if there are many solutions with a high neutral degree. 
A \emph{neutral fitness landscape} can be pictured by a landscape with many plateaus.
A \emph{plateau} $\mathcal{P}$ is a set of pairwise neighboring solutions with the same fitness values:
$ \forall s \in \mathcal{P}, \exists s' \in \mathcal{P}, s \in \V(s')
, f(s^\prime) = f(s)$.
A \textit{portal} in a plateau is a solution that has at least one neighbor with a better fitness value, {\itshape i.e.} a lower fitness value in a minimization context.
Notice that a local optimum can still have some neutral neighbors.
A plateau which contains a local optimum is called a Local Optima Plateau (LOP).

The average or the distribution of neutral degrees over the landscape may be used to test the level of neutrality of a problem. 
This measure plays an important role in the dynamics of local search algorithms \cite{wilke01,verel07}.
When the fitness landscape is neutral, the main features of the landscape can be described by its LOP,
that may be sampled by \textit{neutral walks}.
A neutral walk $W_{neut} = (s_0, s_1, \ldots, s_m)$ from $s$ to $s^\prime$ 
is a sequence of solutions belonging to $S$ where $s_0=s$ and $s_m=s^\prime$
and for all $i \in \{ 0, \dots , m - 1 \}$,
$s_{i+1} \in \V_n(s)$.
Hence, to escape from a LOP, a heuristic method has to perform a neutral walk in order to find a portal.

\subsection{Neutrality in the PFSP}

In scheduling problems, and in particular in flowshop scheduling problems, 
it is well known that several solutions may have the same fitness value, 
\textit{i.e.} the neutral degree of feasible solutions is not null. 
This has been confirmed in a recent work that analyzes the neutrality of Taillard's PFSP instances \cite{Marmion:Lion2011}. 
It has been shown that the neutrality increases when the number of jobs and/or the number of machines increase. 
Moreover,
this study revealed that very few local optima have no neutral neighbor and that the local optima plateaus are numerous and quite large. 
Experiments highlighted that most  random neutral walks on a LOP are able to find a portal,
{\it i.e.} numerous portals exist.
This indicates that exploring a plateau allows to lead to a more interesting solution.
These comments lead us to make proposals on the way to exploit this neutrality
in order to guide the search more efficiently. 
More precisely, we will study how this neutrality can be exploited within an iterated local search algorithm.

\subsection{Iterated Local Search}
\label{sec:ils}

Iterated Local Search (ILS) \cite{lourenco:2002} is a simple and powerful heuristic methodology that applies a local move iteratively  to a single solution.
In order to design an ILS, four elements have to be defined:
($i$) an initialization strategy, 
($ii$) a (basic) local search, 
($iii$) a perturbation, and
($iv$) an acceptance criterion.
The \textit{initialization strategy} generates one solution from the search space. 
The \textit{local search} must lead the current solution to a local optimum.
The \textit{perturbation} process allows to modify a solution in a different way that the neighborhood relation, in order to jump over the landscape. 
The \textit{acceptance criterion} determines whether the local optimum should be kept for further treatments.
Let us remark that both the perturbation process and the acceptance criterion could be based on an history of solutions found during the search.
Thus, during the search,
if the current optimum is not satisfying, the search process is able to restart from a better local optimum, found in a previous iteration.
The search is stopped once a termination condition is satisfied.

The general scheme of ILS, given in \cite{lourenco:2002} is as follows:
($i$) after generating an initial solution, the local search finds a local optimum,
until the termination condition is met, 
($ii$) the current solution is perturbed,
($iii$) the local search is applied to find another local optimum,
($iv$) it is accepted or not regarding the acceptation criterion.
The best solution found along the search is returned as output of the algorithm.

\section{Neutrality-based Iterated Local Search}
\label{sec:nils}

In this section, we propose an ILS based on neutrality, called \textit{Neutrality-based Iterated Local Search} (NILS).
We first give the local search and the perturbation strategy used at every NILS iteration,
and then we discuss the general principles of NILS. 

\subsection{Local Search: First-Improving Hill-Climbing}
\label{sec:FIHC}

NILS is based on the iteration of a First-Improving Hill-Climbing (FIHC) algorithm, where 
the current solution is replaced by the first encountered neighbor that \textit{strictly} improves it.
Each neighbor is explored only once, and the neighborhood is evaluated in a random order.
FIHC stops on a local optimum.

\subsection{Perturbation: Neutral Walk-based Perturbation}
\label{sec:Perturb}

In general, there exists two possible ways of escaping from a local optimum in a neutral fitness landscape:
either performing neutral moves on the Local Optima Plateau (LOP) until finding a portal, 
or performing a kick move which is a `large step' move.
When a neutral move is applied, it is assumed that the \textit{exploitation} of the neutral properties helps to find a better solution.
On the contrary, when a kick move is applied, it supposes that portals are rare,
and that the \textit{exploration} of another part of the search space is more promising.

The proposed Neutral Walk-based Perturbation (NWP), given in Algorithm~\ref{algoPerturb}, 
deals with this tradeoff between exploitation and exploration of LOP.
First, NWP performs a random neutral walk on a LOP.
The maximum number of allowed neutral steps on the LOP is tuned by a parameter denoted by $MNS$ (Maximal Number of Steps).
Along the neutral walk, as soon as a better solution (a portal) is found,
the neutral walk is stopped and the current solution is replaced by this neighboring solution.
Otherwise, if the neutral walk does not find any portal, the solution is kicked.
The kick move corresponds to a large modification of the solution.
Like in FIHC, each neighbor is explored only once, and the neighborhood is evaluated in a random order.

\begin{algorithm}[t]
\caption{Neutral Walk-based Perturbation (NWP)}
\label{algoPerturb}
\begin{algorithmic}
\STATE step $\leftarrow$ 0, better $\leftarrow$ \FALSE
\WHILE{step $<MNS$ \AND \NOT better \AND $|\V_n(s)|>0$}
	\STATE choose $s^{\prime} \in \V (s)$ such that $f(s^{\prime}) \leq f(s)$
	\IF{$f(s^{\prime}) < f(s)$}
		\STATE better $\leftarrow$ \TRUE
	\ENDIF
	\STATE $s \leftarrow s^{\prime}$ 
	\STATE step $\leftarrow$ step$+1$
\ENDWHILE
\IF{\NOT better}
	\STATE $s \leftarrow$ \textbf{kick}($s$)
\ENDIF
\end{algorithmic}
\end{algorithm}

\subsection{NILS: a Neutrality-based ILS}

NILS is based on the FIHC local search and the NWP perturbation scheme. 
Its acceptance criterion always accepts the current solution.
After the initialization of the solution and the first execution of the FIHC,
NILS iterates two phases:
($i$)~a neutral phase which performs neutral moves on LOP, 
($ii$) a strictly improving phase.
The neutrality of the problem is taken into account during phase~($i$). 
Indeed, the neutral walk on the LOP is used to cross a large part of the search space.
When the density of portals on plateaus is low, the $MNS$ parameter avoids to spread on the plateau unnecessarily.
In such a case, a ``restart'' technique is used: the solution is kicked to escape from its neighborhood. 
Phase~($ii$) strictly improves the current solution using the neighborhood operator under consideration.
The neutral phase allows to exploit the neutrality of the search space from the local optima by visiting the corresponding LOP.
Thus, a LOP is considered as a large frontier until a better local optimum.
When a portal is found, the FIHC algorithm is executed in order to find this local optimum.
However, when the frontier is too large, such a local optimum can be difficult to reach.
Thus, the NILS algorithm restarts its search from another part of the search space, expecting that the
next LOP would be easier to cross. 
As a consequence, the tradeoff between exploitation and exploration of the plateaus is directly tuned by the single NILS parameter:
the $MNS$ value.

\section{Neutrality-based Iterated Local Search for the Permutation Flowshop Scheduling Problem}
\label{sec:analysis}

\subsection{Experimental Design}

Experiments are driven using a set of benchmark instances originally proposed by Taillard \cite{taillard:1993} for the flowshop scheduling problem, and widely used in the literature \cite{stutzle:1998,ruiz2005}.
We investigate different values for the number of machines $M \in \{ 5,10,20 \}$ and for the number of jobs $N \in \{ 20,50,100,200,500 \}$.
The processing time $t_{ij}$ of job $i \in N$ and machine $j \in M$ is an integer value generated randomly, according to a uniform distribution $\mathcal{U}(0,99)$. 
For each problem size ($N \times M$), 10 instances are available.
Note that, as mentioned on the Taillard's website\footnote{\url{http://mistic.heig-vd.ch/taillard/problemes.dir/ordonnancement.dir/ordonnancement.html}},
very few instances with $20$ machines have been solved to optimality.
For $5$- and $10$-machine instances, optimal solutions have been found, requiring for some of them a very long computational time.
Hence, the number of machines seems to be very determinant in the problem difficulty.
That is the reason why the results of the paper will be exposed separately for each number of machines.

The performances of the NILS algorithm are experimented on the first Taillard's instance of each size,
given by the number of jobs ($N$) and the number of machines ($M$).
For each size, several $MNS$ values are investigated:
\begin{itemize}
\item For $N=20$, $~~MNS \in \{0,10,20,50,100\}$ 
\item For $N=50$, $~~MNS \in \{0,10,50,70,100,150,300,900,1800\}$ 
\item For $N=100$, $MNS \in \{0,50,100,300,1200,5000,10000,20000,40000\}$ 
\item For $N=200$, $MNS \in \{0,100,200,500,1000,4000,8000,30000,60000\}$ 
\item For $N=500$, $MNS \in \{0,250,500,1000,5000,10000,15000,20000,40000\}$ 
\end{itemize}
NILS is tested $30$ times for each $MNS$ value and each problem size.
Let us remark that when $MNS=0$, NILS corresponds to an ILS algorithm without any neutrality consideration. 
Hence, we are able to compare an ILS that uses neutrality to a classical one.
As the distribution of the $30$ final fitness values is not symmetrical, 
the average and the standard deviation are less statistically meaningful. 
Therefore, for each problem size and each $MNS$ value, the median and the boxplot of the 30 fitness values of solutions found are presented.
For the other statistical quantities, the average value and the standard deviation of the 30 executions are computed.

In order to let any configuration enough time to converge, the NILS termination condition is given in terms of a maximal number of evaluations,
set to $2.10^7$.
The insertion operator (Section \ref{subsec:nbh}) defines the neighborhood used in the first improving hill-climbing (FIHC) and in the random neutral walk (NWP).
In the NILS perturbation scheme,
the solution is kicked when the neutral walk does not find any portal. In this study, the kick move corresponds to 3 randomly chosen exchange moves.
As empirically shown in \cite{ruiz2005}, this kick is instance-dependent.
However, as our work does not attempt to set the best parameters for the kick move,
we choose a proper value that obtained reasonably good performances for all instances.

\subsection{Experimental Results}
\label{sectionPerf}

\begin{table}[t]
\caption{NILS performances according to the number of jobs ($N$) and to the number of machines ($M$). 
The best known solution from the literature is given. 
NILS$_*$ gives the best performance found for all $MNS$ parameter values.
The median of the 30 fitness values is calculated for each $MNS$ value.
NILS$_0$ gives the median for $MNS=0$.
NILS$_{max}$ gives the median for the maximal tested $MNS$ value.
The best value between NILS$_0$ and NILS$_{max}$ is given in bold type.}
\label{tabPerf}
\begin{small}
\begin{center}
\begin{tabular}{ccccp{0.4cm}cc}

   & & $~~$Literature$~~$ & $~~$ NILS$_{*}$ $~~$ & & $~~$ NILS$_0$ $~~$ & $~~$ NILS$_{max}$ $~~$  \\ 
     \cline{3-4} \cline{6-7}
    $~~M$ $~~$ & $~~N$ $~~$ &  \multicolumn{2}{c}{best} & &  \multicolumn{2}{c}{median} \\
  \hline \\ 
    $~$5 &  $~$20 & $~$1278 & $~$1278 & & $~~~$1278 &  $~~~~$1278\\
    &  $~$50 & $~$2724 & $~$2724 & & $~~~$2724 &  $~~~~$2724\\  
    & 100 & $~$5493 & $~$5493 & & $~~~$5493 &  $~~~~$5493\\
  \\
 10 &  $~$20 &  $~$1582 &  $~$1582 & & $~~~$1582 &  $~~~~$1582\\
    &  $~$50 &  $~$2991 &  $~$3025 & & $~~~$3042 & \textbf{ $~~~$3025} \\  
    & 100 &  $~$5770 &  $~$5770 & & $~~~$5784 &  \textbf{$~~~~$5771} \\
    & 200 & 10862 & 10872 & & $~~$10885 &  \textbf{$~~$10872}\\
  \\
 20 &  $~$20 &  $~$2297 &  $~$2297 & & $~~~$2297  &  $~~~~~$2297 \\
    &  $~$50 &  $~$3847 &  $~$3894 & & $~~~$3925 &  \textbf{$~~~~$3917} \\  
    & 100 &  $~$6202 &  $~$6271 & & $~$6375.5 & \textbf{$~~$6315.5}  \\
    & 200 & 11181 & 11303 & & 11405.5 & \textbf{11334.5} \\
    & 500 & 26059 & 26235 & & 26401.5 & \textbf{26335.5} 
\end{tabular}
\end{center}
\end{small}
\end{table}

\begin{figure}[t]
\begin{minipage}[t]{0.30\linewidth}
\centering
\includegraphics[scale=0.19]{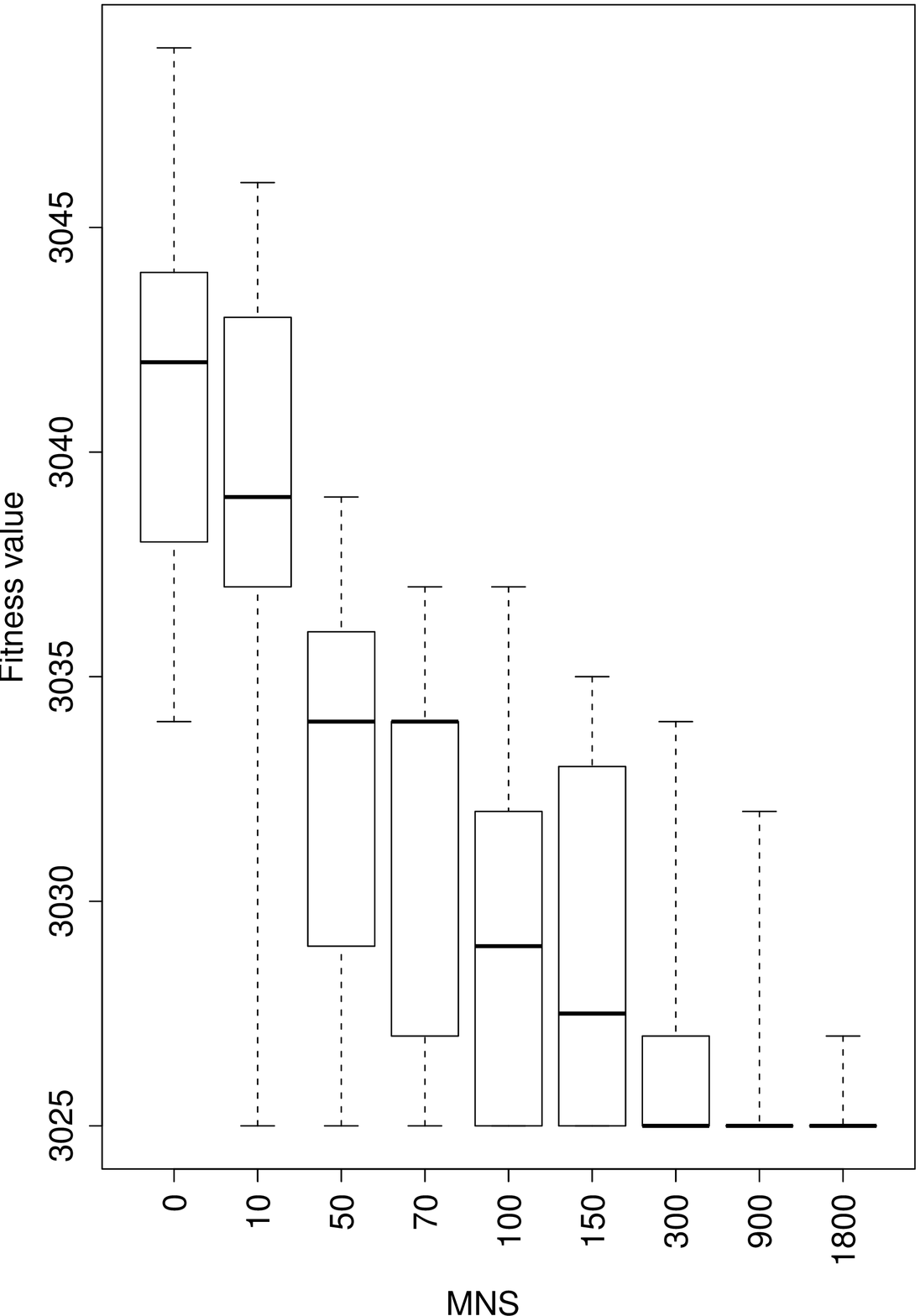}
\label{fit5010}
\end{minipage} \hfill
\begin{minipage}[t]{0.30\linewidth}
\centering
\includegraphics[scale=0.19]{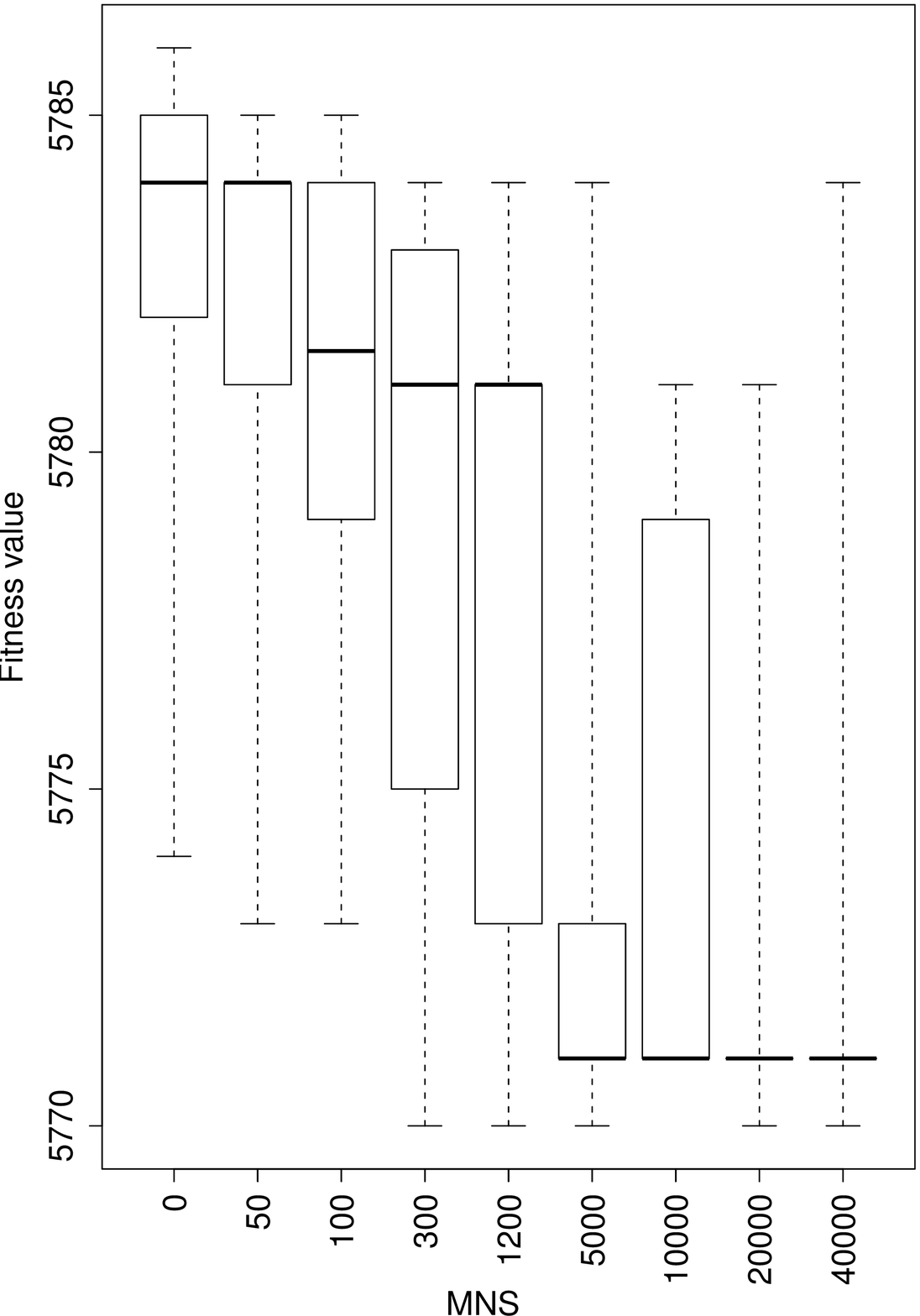}
\label{fit10010}
\end{minipage} \hfill
\begin{minipage}[t]{0.30\linewidth}
\centering
\includegraphics[scale=0.19]{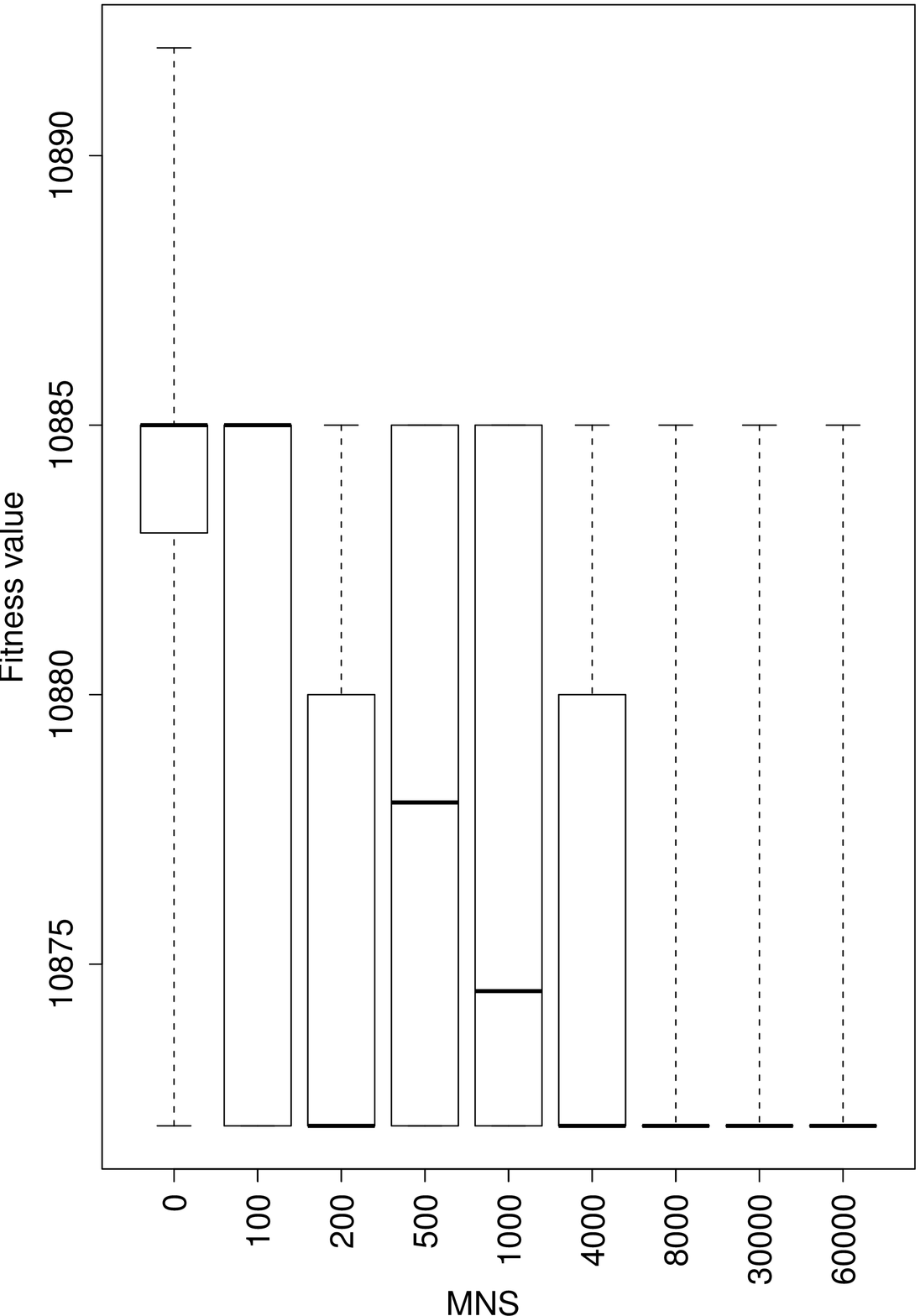}
\label{fit20010}
\end{minipage} \\
\begin{minipage}[t]{0.30\linewidth}
\centering
\includegraphics[scale=0.19]{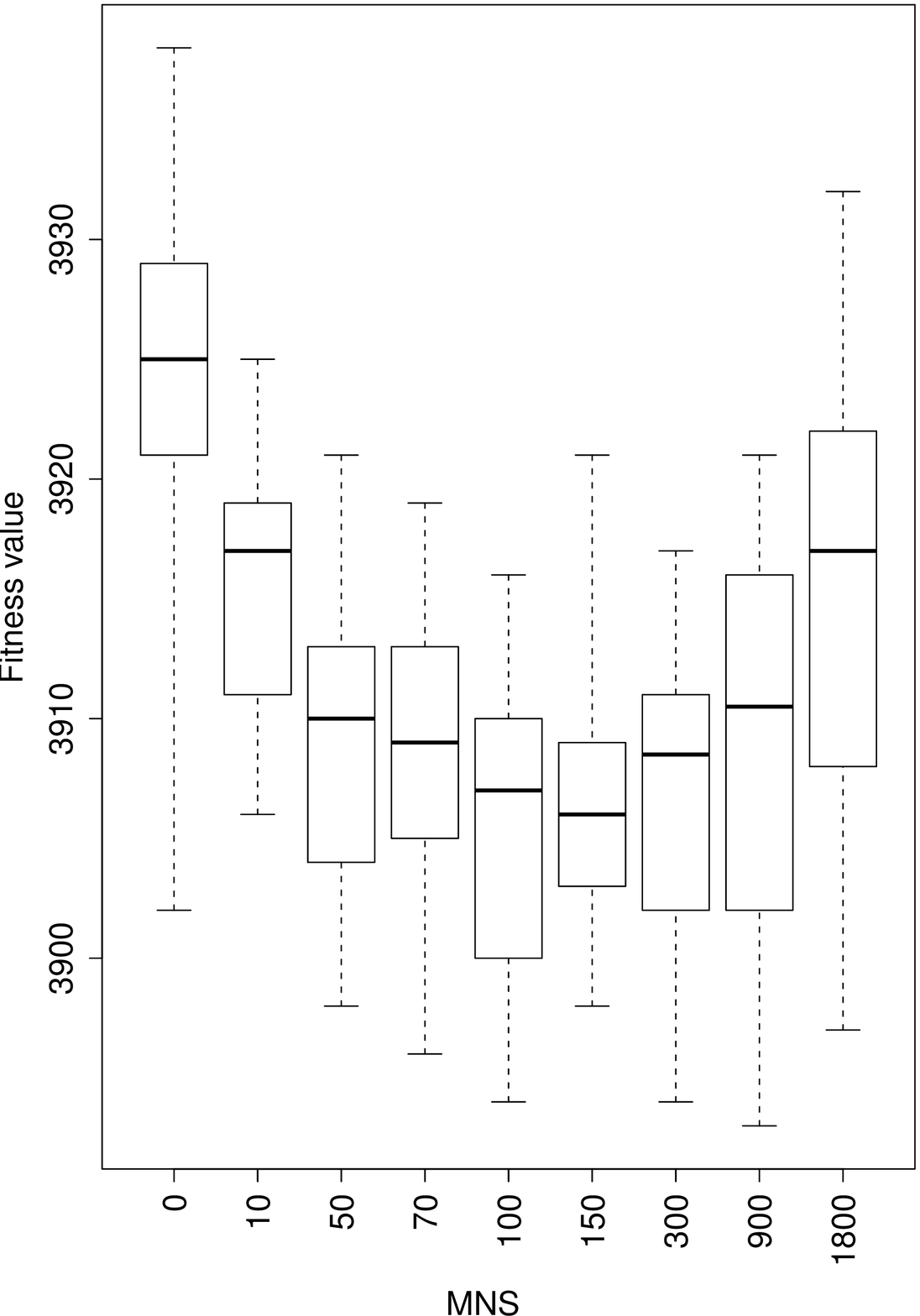}
\label{fit5020}
\end{minipage} \hfill
\begin{minipage}[t]{0.30\linewidth}
\centering
\includegraphics[scale=0.19]{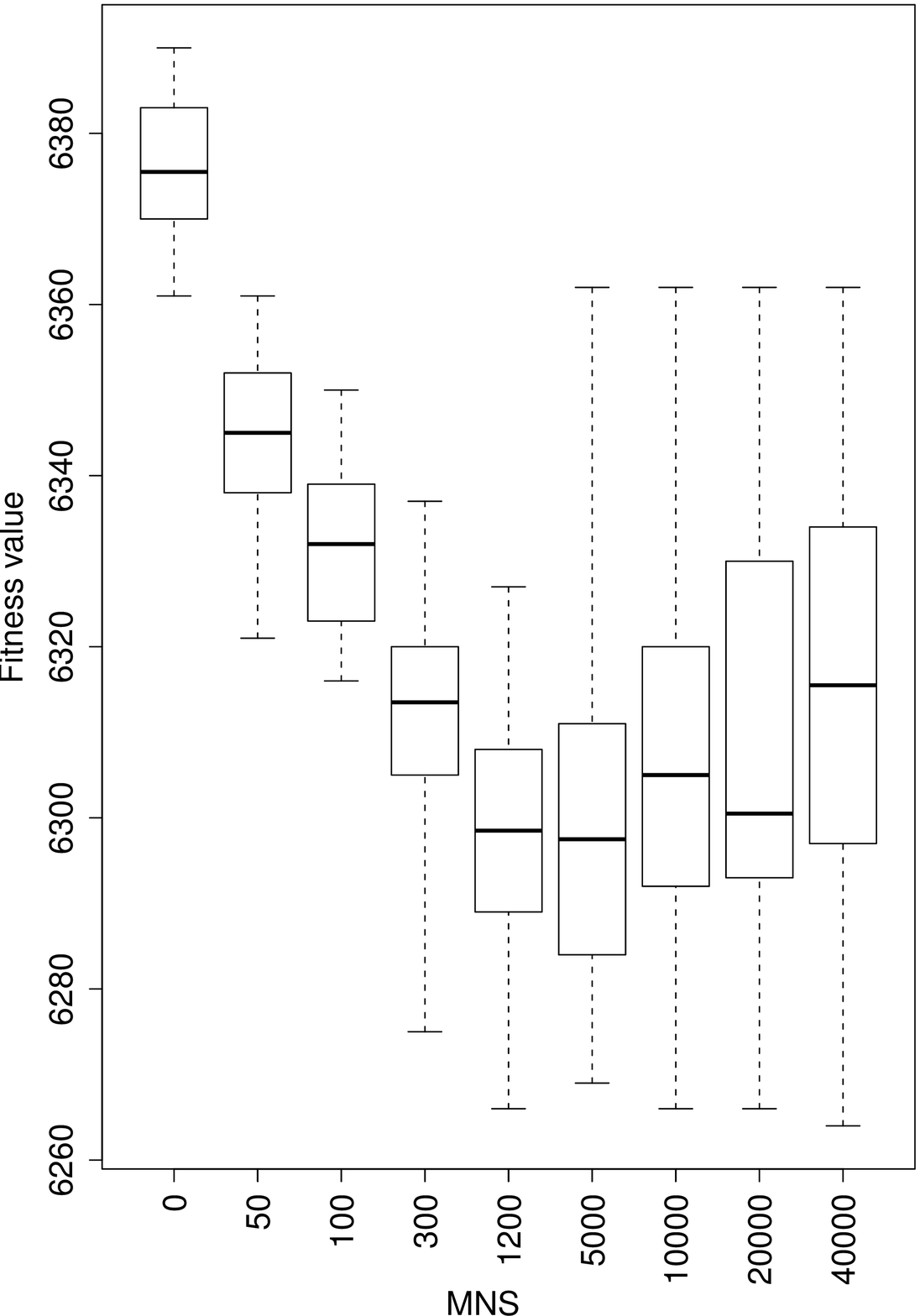}
\label{fit10020}
\end{minipage} \hfill
\begin{minipage}[t]{0.30\linewidth}
\centering
\includegraphics[scale=0.19]{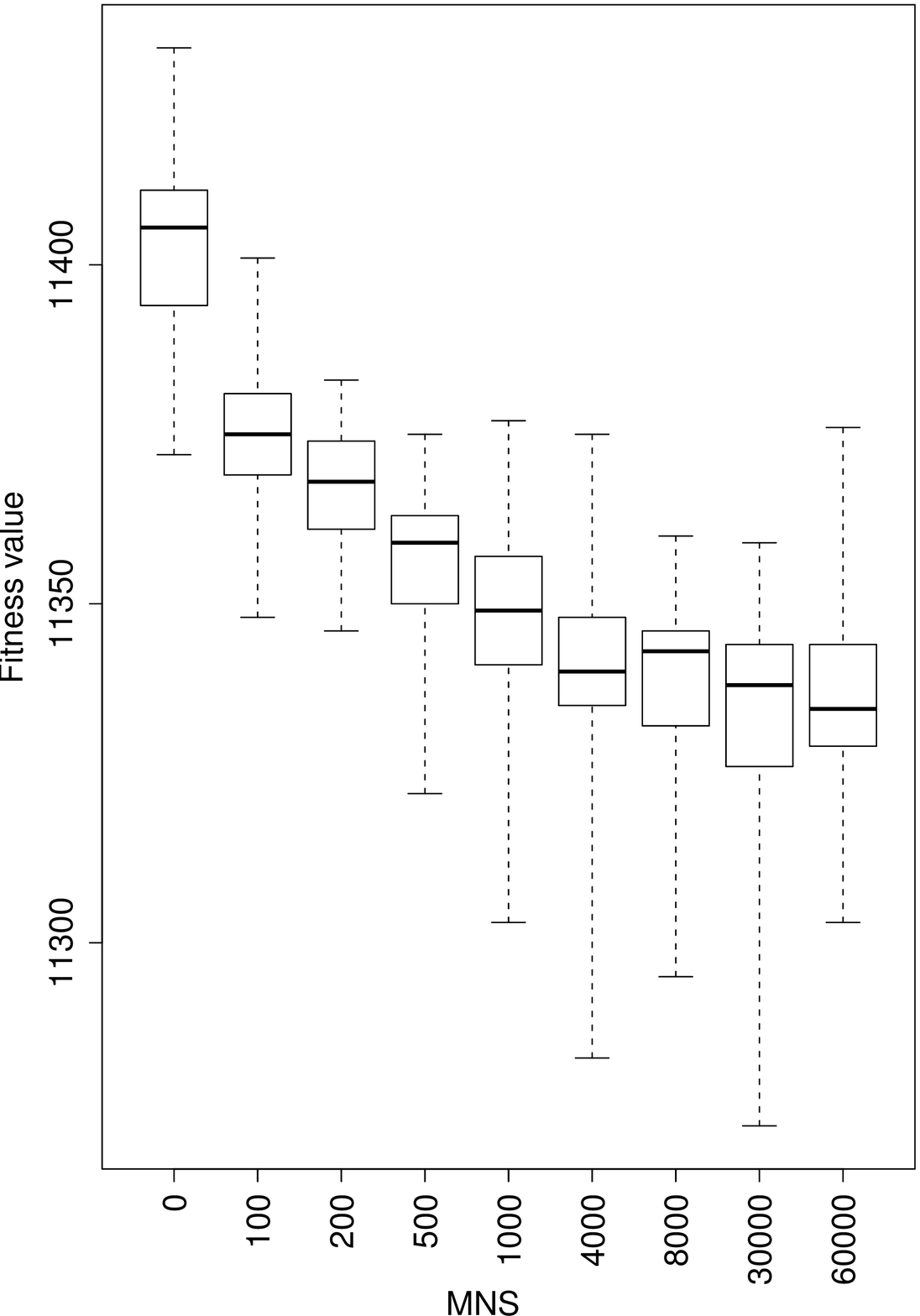}
\label{fit20020}
\end{minipage}
\caption{Boxplots of fitness values of the 30 solutions found after $2.10^7$ evaluations. The boxplots for $10$-machines and $20$-machines instances are represented top and bottom respectively and the $50$-,$100$-,$200$-jobs instances from left to right.}
\label{boxplotFit}
\end{figure}

This section examines the benefit of taking neutrality into account in the search process.
We denote by NILS$_{x}$, the NILS algorithm with the following parameter setting: $MNS=x$. 
Results obtained with $MNS=0$ and $MNS>0$ are compared with each other.
Table \ref{tabPerf} presents, for each problem size ($M$ $\times$ $N$), the fitness value of the best solution found in comparison to the best known solution from the literature.
Let us remind that a minimization problem is under consideration.
The median of the 30 fitness values found by NILS$_{0}$
and NILS with the larger $MNS$ value (NILS$_{max}$) are also given.
For $5$-machine instances, the best solution is reached by every NILS configuration.
This means that using the neutral walk is as interesting as the classical ILS model.
The same conclusion can be made for all $20$-job instances.
For the remaining instances, the performance is quite far from the best known results from the literature. 
However, the median value of NILS$_{max}$ is better than the one of NILS$_{0}$.
For example, for $N=200$ and $M=20$,
NILS$_{max}$ clearly outperforms NILS$_{0}$.
The best results are reached when neutral walks on plateaus are allowed,
which validates the use of neutrality.

When the results are compared with the degree of neutrality,  
we can give additional conclusions.
For $20$-machine instances, the average degree of neutrality of solutions is quite low (around $5 \%$ in average) \cite{Marmion:Lion2011}.
However, the performance of NILS is better.
This could be surprising at first sight.
In fact, for those instances,
the degree of neutrality is sufficient for the neutral random walks to cross a wide search space with the same fitness level. 
For the instances with a higher degree of neutrality,
the density of portals on the plateaus decreases, 
and a pure random walk on a plateau is less efficient to find a portal quickly.

\subsection{Influence of the $MNS$ Parameter}

\begin{figure}[t]
\begin{center}
\begin{tabular}{cc}
\includegraphics[scale=0.19]{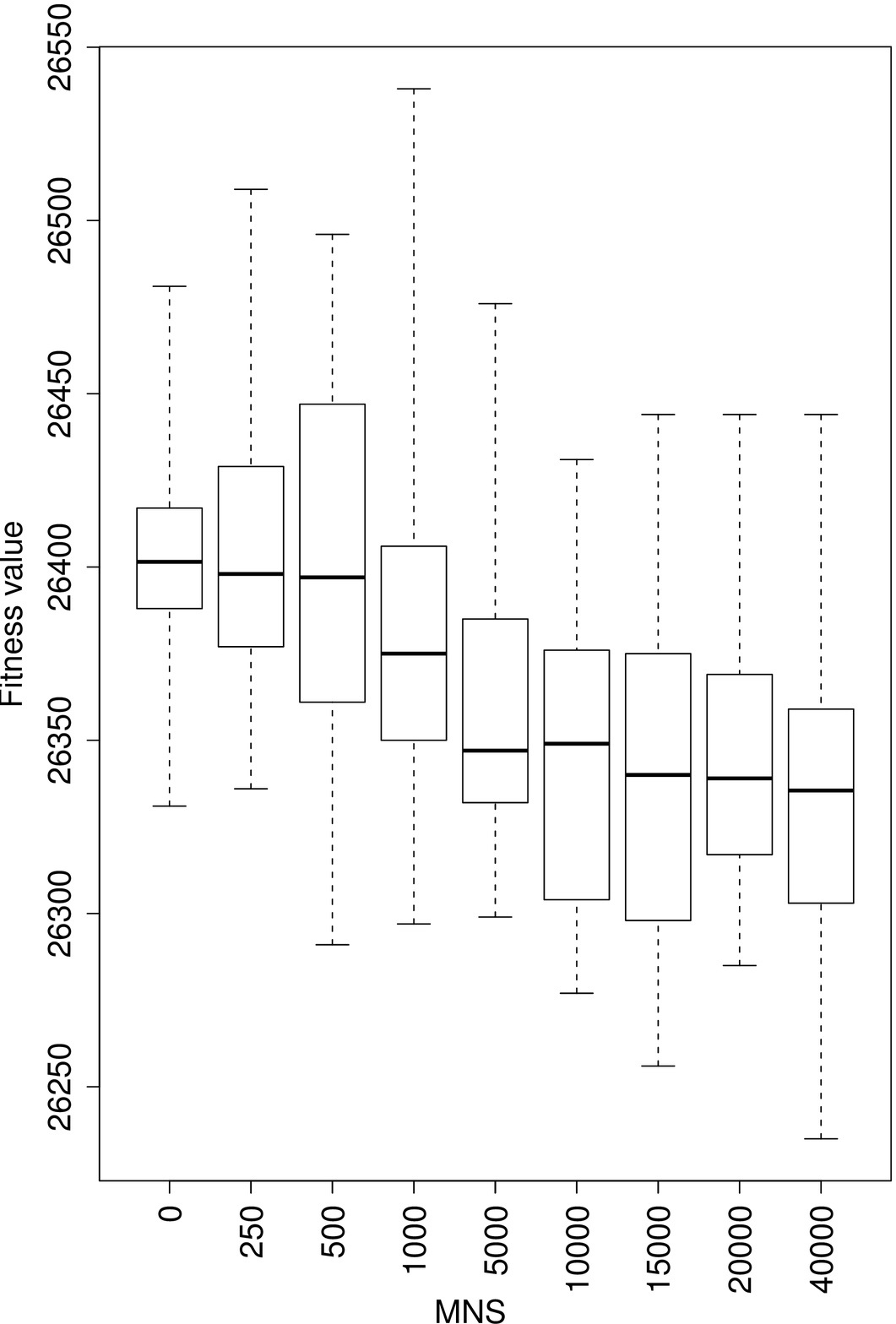} & \includegraphics[scale=0.50]{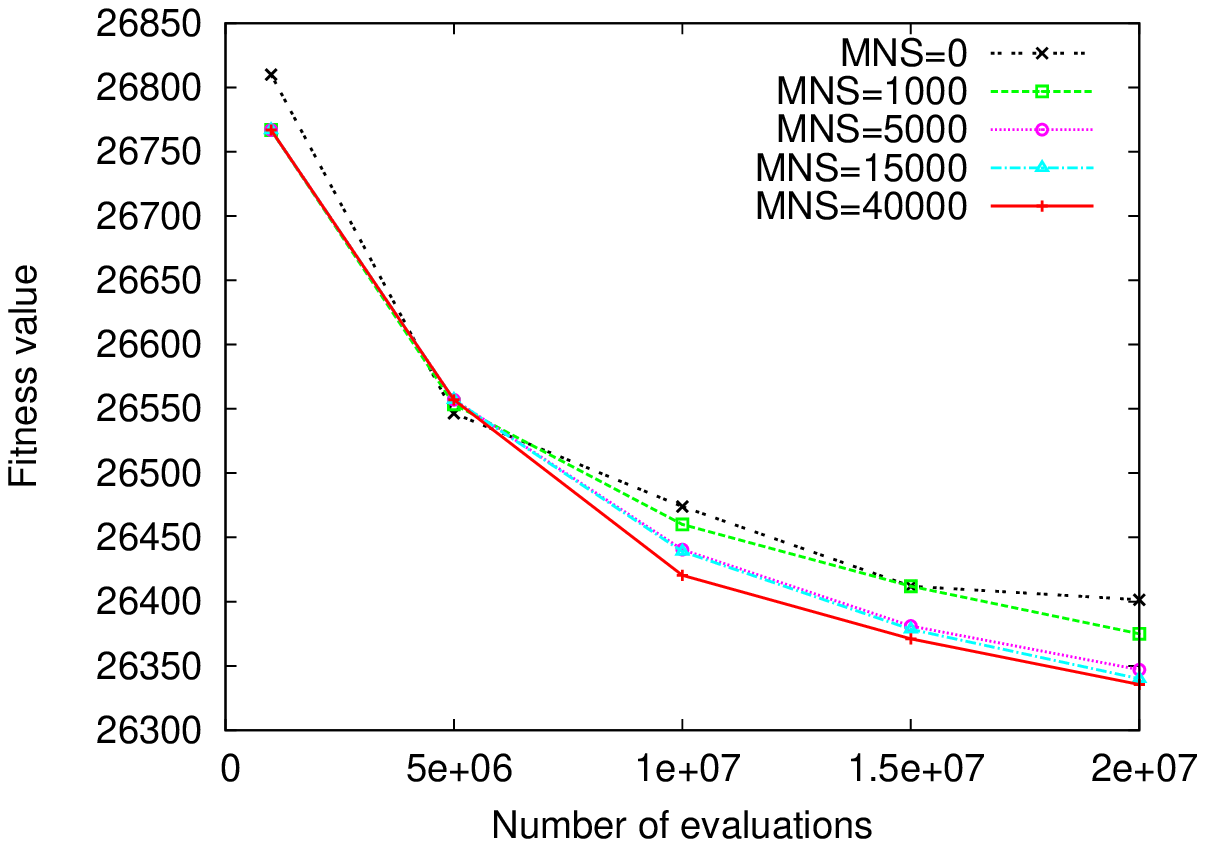} \\
(a) & (b) \\
\includegraphics[scale=0.42]{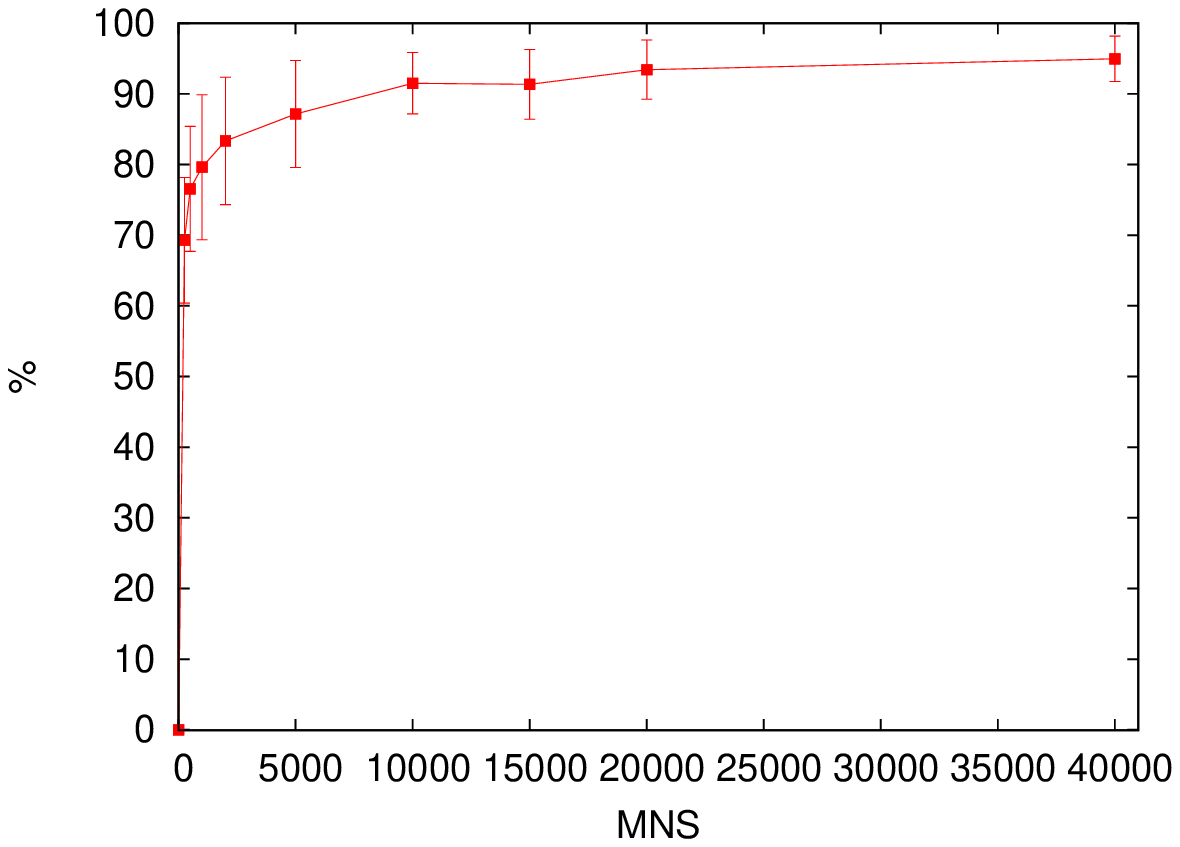} & \includegraphics[scale=0.42]{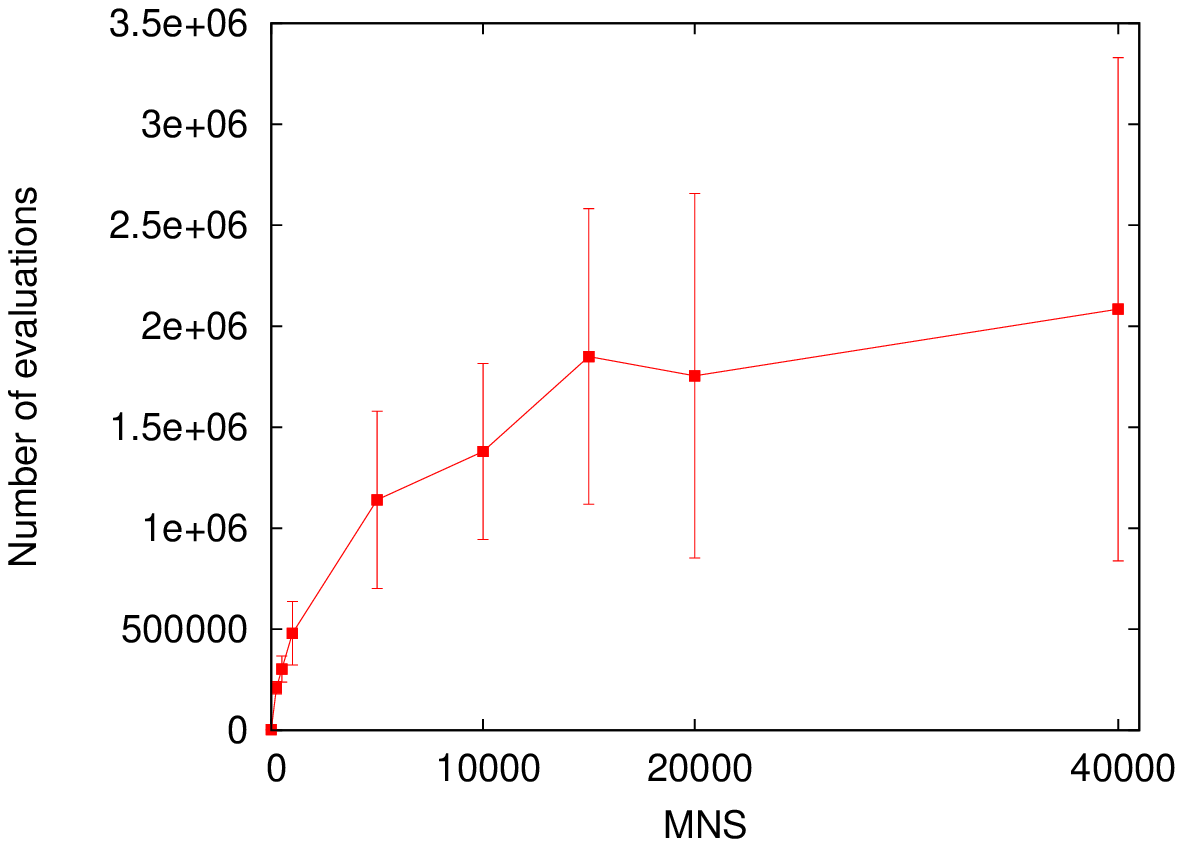} \\
(c) & (d) 
\end{tabular}
\caption{Results for N=500 and M=20. Boxplots of fitness values of the 30 solutions found after $2.10^7$ evaluations (a). Median performance according to the number of evaluations (b).
Average percentage (and standard deviation) of visited local optima plateaus where the perturbation led to a portal (c). 
Number of evaluations (average number and standard deviation) lost during the perturbation step according to the $MNS$ values (d).}
\label{result50020}
\end{center}
\end{figure}

Since random neutral walks on a local optima plateau seem to be efficient in finding improving solutions,
we can wonder whether the neutral walk should be large or not. 
The $MNS$ parameter corresponds to the maximal number of neutral steps allowed to move on the LOP.
In this section,
we study the performance using different $MNS$ values that depend on the number of jobs and so, that is related to the neighborhood size.
Figure \ref{boxplotFit} shows the boxplot of fitness values for instances up to $50$ jobs and $10$ machines.
For $10$-machine instances~(top),
the median fitness value starts by decreasing with the $MNS$ value, and then stabilizes:
the results are equivalent for large $MNS$ values.
As they are not distributed normally, the Mann-Whitney statistical test was used to check this tendency. 
The pairwise tests validate that the results are not statistically different for highest $MNS$ values.
Increasing the value of the $MNS$ parameter does not deteriorate the performance.
Therefore, it appears that this parameter can be set to a large value.
For $20$-machine instances (bottom), best performances are always found by a NILS algorithm that exploit the LOP.
The same happens for $N=200$ and $N=500$ (Figure~\ref{result50020} (a)).
However, for $N=50$ and $N=100$, the performance decreases with the $MNS$ value.
This remark is confirmed by the Mann-Whitney statistical test.
Therefore, in some cases, an over-exploitation of the plateaus can become too costly.
Here, we see that the tradeoff between neutrality exploitation and search space exploration should be considered carefully
as it seems to exist a more-suited $MNS$ value for some instance sizes.

The performances after $2.10^7$ evaluations are better with the maximum number of neutral steps.
Figure \ref{result50020} (b) represents the median of the 30 fitness values according to different numbers of evaluations for $N=500$ and $M=20$.
For each possible number of evaluations, the median performance is always higher when the $MNS$ value is large.
Moreover, the larger the problem size, the better the fitness value found using neutrality along the search (see Figure \ref{result50020} (b)). 
Whatever the fitness level, from $26401.5$ to $26335.5$, using neutrality clearly improves the performance.


\subsection{Benefits and Costs of Neutral Moves}
\label{sectionStepMaxNW}

In order to deeply analyze the tradeoff between exploitation and exploration, tuned by the $MNS$ parameter,
we here estimate the benefit and the cost of neutral moves on the plateaus.
Figure~\ref{result50020}(c) computes the number of portals reached by the neutral walks for different $MNS$ values
for $N=500$ and $M=20$.
When the $MNS$ value increases, the number of reached portals increases, even if the slope decreases.
This indicates that the few additional portals reached with a larger $MNS$ value are interesting ones, since they lead to better solutions
(Figure~\ref{boxplotFit}).

The cost of the neutral moves is the total number of evaluations when the neutral walk fails to find a portal.
Figure \ref{result50020} (d) gives this number for $N=500$ and $M=20$.
Even if this cost increases with the $MNS$ value, we can notice that the slope decreases.
This indicates that a larger neutral walk keeps interesting performance.
Nevertheless, the slope of the cost is higher than the one of the benefit. 
It suggests that a tradeoff value can be reached in favor of a kick-move exploration for very large $MNS$ values.

\section{Conclusion and Future Works}

In this paper, we exploited the property of neutrality in order to design an efficient local search algorithm.
Indeed, numerous combinatorial optimization problems contain a significant number of equivalent neighboring solutions.
A Neutrality-based Iterated Local Search (NILS), that allows neutral walks to move on the plateaus, is proposed.
The performance of the NILS algorithm has been experimented on the permutation flowshop scheduling problem
by regarding the maximum length of neutral walks.
As revealed in a previous study~\cite{Marmion:Lion2011},
the fitness landscape associated with this problem has a high degree of neutrality.
Our experimental analysis shows that neutral walks allow to find improving solutions,
and that the longer the neutral walk, the better the solution found.
NILS is able to take advantage of applying neutral moves on plateaus, without any prohibitive additional cost.
This can be explained by the relatively high density of portals over plateaus which can be reached by a random neutral walk.

In their iterated greedy algorithm proposed to solve the flowshop problem investigated in the paper,
Ruiz et al. \cite{ruiz_stutzle:2006} combine a steepest descent local search with 
a destruction and a construction phases for perturbation, and an acceptance criterion based on a temperature.   
Let us remark that the authors suggest that the temperature level has not a significant influence on the overall performance.
However, for the problem at hand, a solution has a non negligible number of neutral neighbors and the local optima plateaus are quite large.
Thus, the destruction and construction phases can easily lead to a solution with the same fitness value.
This perturbation can be compared to the NILS perturbation between a random neutral walk and a kick-move exploration.
In the NILS algorithm, this tradeoff is explicitly tuned by a single parameter.

Considering those remarks together with the neutral exploitation of NILS,
it suggests that the good performance of Ruiz et al.'s algorithm \cite{ruiz_stutzle:2006} is probably due to
the degree of neutrality of the flowshop scheduling problem.
NILS uses this property more explicitly, with a lower implementation cost and less parameters.

This first version of the NILS algorithm explores the neighborhood in a random order, 
and both the components of hill-climbing and of random neutral walk are based on a first-improving strategy.
Future works will be dedicated to design efficient guiding methods based on the evolvability of solutions,
in order to reach the portals more quickly.
Moreover, a similar work will allow to better understand the dynamics of the NILS algorithm on other combinatorial optimization problems
where a high degree of neutrality arises.
Hopefully, this will allow us to propose an adaptive control of the single parameter of the NILS algorithm:
the maximum length of the neutral walk.

%
%
\bibliographystyle{splncs}


\end{document}